\documentclass{article}

\usepackage{microtype}
\usepackage{graphicx}
\usepackage{subfigure}
\usepackage{booktabs} % for professional tables

\usepackage{hyperref}

\usepackage[accepted]{icml2018}

\usepackage[utf8]{inputenc} % allow utf-8 input
\usepackage[T1]{fontenc}    % use 8-bit T1 fonts
\usepackage{hyperref}       % hyperlinks
\usepackage{url}            % simple URL typesetting
\usepackage{booktabs}       % professional-quality tables
\usepackage{amsfonts}       % blackboard math symbols
\usepackage{nicefrac}       % compact symbols for 1/2, etc.
\usepackage{microtype}      % microtypography
\usepackage{amsmath,amssymb}
\usepackage{mathtools}
\usepackage{enumerate}

\icmltitlerunning{Double Uncertain Exploration}

\begin{document}

\twocolumn[
\icmltitle{The Potential of the Return Distribution for Exploration in RL}

% It is OKAY to include author information, even for blind
% submissions: the style file will automatically remove it for you
% unless you've provided the [accepted] option to the icml2018
% package.

% List of affiliations: The first argument should be a (short)
% identifier you will use later to specify author affiliations
% Academic affiliations should list Department, University, City, Region, Country
% Industry affiliations should list Company, City, Region, Country

% You can specify symbols, otherwise they are numbered in order.
% Ideally, you should not use this facility. Affiliations will be numbered
% in order of appearance and this is the preferred way.
\icmlsetsymbol{equal}{*}

\begin{icmlauthorlist}
\icmlauthor{Thomas M. Moerland}{to}
\icmlauthor{Joost Broekens}{to}
\icmlauthor{Catholijn M. Jonker}{to}
\end{icmlauthorlist}

\icmlaffiliation{to}{Department of Computer Science, Delft University of Technology, The Netherlands}

\icmlcorrespondingauthor{Thomas M. Moerland}{T.M.Moerland@tudelft.nl}

% You may provide any keywords that you
% find helpful for describing your paper; these are used to populate
% the "keywords" metadata in the PDF but will not be shown in the document
\icmlkeywords{reinforcement learning, exploration, return distribution, uncertainty, variance, deep learning}

\vskip 0.3in
]

% this must go after the closing bracket ] following \twocolumn[ ...

% This command actually creates the footnote in the first column
% listing the affiliations and the copyright notice.
% The command takes one argument, which is text to display at the start of the footnote.
% The \icmlEqualContribution command is standard text for equal contribution.
% Remove it (just {}) if you do not need this facility.

\printAffiliationsAndNotice{}  % leave blank if no need to mention equal contribution
%\printAffiliationsAndNotice{\icmlEqualContribution} % otherwise use the standard text.

\begin{abstract}
This paper studies the potential of the return distribution for exploration in deterministic reinforcement learning (RL) environments. We study network losses and propagation mechanisms for Gaussian, Categorical and Gaussian mixture distributions. Combined with exploration policies that leverage this return distribution, we solve, for example, a randomized Chain task of length 100, which has not been reported before when learning with neural networks.
\end{abstract} 

\section{Introduction}
Reinforcement learning (RL) is the dominant class of algorithms to learn sequential decision-making from data. Most RL approaches focus on learning the mean action value $Q(s,a)$. Recently, \citet{bellemare2017distributional} studied distributional RL, where one propagates the entire return distribution $p(Z|s,a)$ (of which $Q(s,a)$ is the expectation) through the Bellman equation. \citet{bellemare2017distributional} show increased performance in a variety of RL tasks.

However, \citet{bellemare2017distributional} did not yet leverage the return distribution for exploration. In the present paper, we identify the potential of the return distribution for informed exploration. The return distribution may be induced by two sources of stochasticity: 1) our stochastic policy and 2) a stochastic environment. For this work we assume a \textit{deterministic} environment, which makes the return distribution entirely induced by the stochastic policy. Thereby, we may actually act optimistically with respect to this distribution.\footnote{In Section \ref{future}, we more thoroughly discuss the different types of uncertainty present in sequential decision making.} The present paper explores this idea, in the context of neural networks and for different propagation distributions (Gaussian, Categorical and Gaussian mixture). Results show vastly improved learning in a challenging exploration task, which had not been solved with neural networks before. We also provide extensive visual illustration of the process of return-based exploration, which shows a natural shift from exploration to exploitation.

\section{Distributional Reinforcement Learning} \label{distr_rl}
We adopt a Markov Decision Process (MDP) \citep{sutton1998reinforcement} given by the tuple $\{\mathcal{S},\mathcal{A},P,R,\gamma\}$. For this work, we assume a discrete action space and deterministic transition and reward functions. At every time-step $t$ we observe a state $s_t \in \mathcal{S}$ and pick an action $a_t \in \mathcal{A} = \{1,2,..,|\mathcal{A}| \} $. The MDP follows the transition dynamics $s_{t+1} = P(s_t,a_t) \in \mathcal{S}$ and returns rewards $r_t(s,a) = R(s_t,a_t) \in \mathbb{R}$. We act in the MDP according to a stochastic policy $\pi(\cdot|s) \in \mathcal{P}(\mathcal{A})$. The (discounted) return $Z^\pi(s,a)$ from a state-action pair $(s,a)$ is a {\it random process} given by

\begin{align} 
&Z^\pi(s,a) = \sum_{t=0}^{\infty} \gamma^t r_t 
\label{eq_return}
\end{align}

where $s_{t+1} = P(s_t,a_t), a_{t+1} \sim \pi(\cdot|s_{t+1}), s_0=s, a_0=a$. The return $Z^\pi$ is a random variable, where the distribution of $Z^\pi$ is induced by the stochastic policy (as we assume a deterministic environment). Eq. \ref{eq_return} can be unwritten in recursive form, known as the {\it distributional Bellman equation} \citep{bellemare2017distributional} (omitting the $\pi$ superscript from now on):

\begin{equation} 
Z(s,a) \overset{\mathrm{d}}{=} r(s,a) + \gamma \mathrm{E}_{A'} [ Z(s',A') ] \label{eq_distr_bellman}
\end{equation}

where $\overset{\mathrm{d}}{=}$ denotes distributional equality \citep{engel2005reinforcement}. The state action value $Q(s,a) = \mathbb{E}_Z [Z(s,a)]$ is the expectation of the return distribution. Applying this expectation to Eq. \ref{eq_distr_bellman} gives

\begin{equation}
Q(s,a) = r(s,a) + \gamma \mathrm{E}_{A' \sim \pi(\cdot|s')} [Q(s',A') ], \label{eq_bellman}
\end{equation}

$s' = P(\cdot|s,a)$, which is known as the Bellman equation \citep{sutton1998reinforcement}. Most RL algorithms learn this mean action value $Q(s,a)$, and explore by some random perturbation of these means. 

\section{Distributional Perspective on Exploration} \label{valuefunctions}
As mentioned in the introduction, the return distribution may be induced by two  sources of stochasticity: 1) our stochastic policy and 2) a stochastic environment. Therefore, if we assume a deterministic environment, then the return distribution is entirely induced by our own policy. As we may modify our policy, it actually makes sense to act optimistically with respect to the return distribution. 

As an illustration, consider a state-action pair with particular mean value estimate $Q(s,a)$. It matters whether this average originates from a highly varying return or from consistently the same return. It matters because our policy may {\it influence} the shape of this distribution, i.e. for the highly varying returns we may actively transform the distribution towards the good returns. In other words, what we really care about in deterministic domains is the best return, or the upper end of the return distribution, because it is an indication of what we may achieve once we have figured out how to act in the future. By starting from broad distribution initializations that gradually narrow when subpolicies converge, we observe a natural shift from exploration to exploitation.

\section{Distributional Policy Evaluation} \label{return_propagation}
Following \citet{bellemare2017distributional}, we introduce a neural network to model the return distribution $p_\phi(Z|s,a)$. For this work we will consider three parametric distributions $p_\phi(Z)$ to approximate the return distribution: Gaussian, Categorical (as previously studied by \citet{bellemare2017distributional}), and Gaussian mixture. 

To perform policy evaluation, we need to discuss two topics: 
\begin{enumerate}
\item How to propagate the distribution through the Bellman equation (based on newly observed data). We will denote the propagated distribution as $q(Z)$.
\item A loss function between the current network predictions and the new target: $\mathrm{L}(p_\phi(Z),q(z))$. 
\end{enumerate}
Due to space restrictions, we will only show the propagation and loss for the Gaussian case. For the Categorical and Gaussian mixture outcome we specify the distributional loss in Appendix \ref{app_losses} and the Bellman propagation in Appendix \ref{app_bellman}. 

\paragraph{Distribution propagation} Define the {\it distributional Bellman back-up operator} $\mathcal{T}$ as the recursive application of Eq. \ref{eq_distr_bellman} to $Z(s,a)$. For a Gaussian network output, $p_\phi(Z|s,a) = \mathcal{N}(Z|\mu_\phi(s,a),\sigma_\phi(s,a))$, we need to propagate both the mean and variance through the Bellman equation.

\begin{align}
\mu^q(s,a) &= \mathrm{E}_Z \Big[ \mathcal{T} Z(s,a) \Big] \nonumber \\
&= \mathrm{E}_Z \Big[ r(s,a) + \gamma \mathrm{E}_{A'} Z(s',A') \Big] \nonumber \\
&= r(s,a) + \gamma \mathrm{E}_{A'} \mu_\phi(s',A')
\end{align}

\vspace{-0.4cm}
\begin{align}
\sigma^q(s,a) &= \mathrm{Sd} \Big[ \mathcal{T} Z(s,a) \Big] = \mathrm{Sd}\Big[ r(s,a) + \gamma \mathrm{E}_{A'} Z(s',A')\Big] \nonumber \\
&= \gamma \mathrm{Sd}\Big[ \mathrm{E}_{A'} Z(s',A')\Big] = \gamma \mathrm{E}_{A'} \sigma_\phi(s',A')
\end{align}

because $\gamma \geq 0$, $\pi(a'|s') \geq 0$, and we assume the next state distributions are independent so we may ignore the covariances.\footnote{For random variables $X,Y$ and scalar constants $a,b,c$ we have: $\mathrm{E}[a + b X + c Y] = a + b \mathrm{E}[X] + c \mathrm{E}[Y]$ and $\mathrm{Var}[a + b X + c Y] = b^2 \hspace{0.1cm} \mathrm{Var}[X] + c^2 \hspace{0.1cm} \mathrm{Var}[Y] + 2bc \hspace{0.1cm} \mathrm{Cov}[X,Y]$.}

In practice, we approximate the expectation over the policy probabilities at the next state $a'$ by sampling a next state once (either on- or off-policy). This is the most common solution in RL, and will be right in expectation over multiple traces. The 1-step bootstrap distribution estimate then becomes $q(Z|s,a) = \mathcal{N}(Z|r(s,a) + \gamma \mu_{\phi}(s',a'),\gamma \sigma_{\phi}(s',a'))$.

\begin{figure*}[t]
  \centering
      \includegraphics[width = 0.94\textwidth]{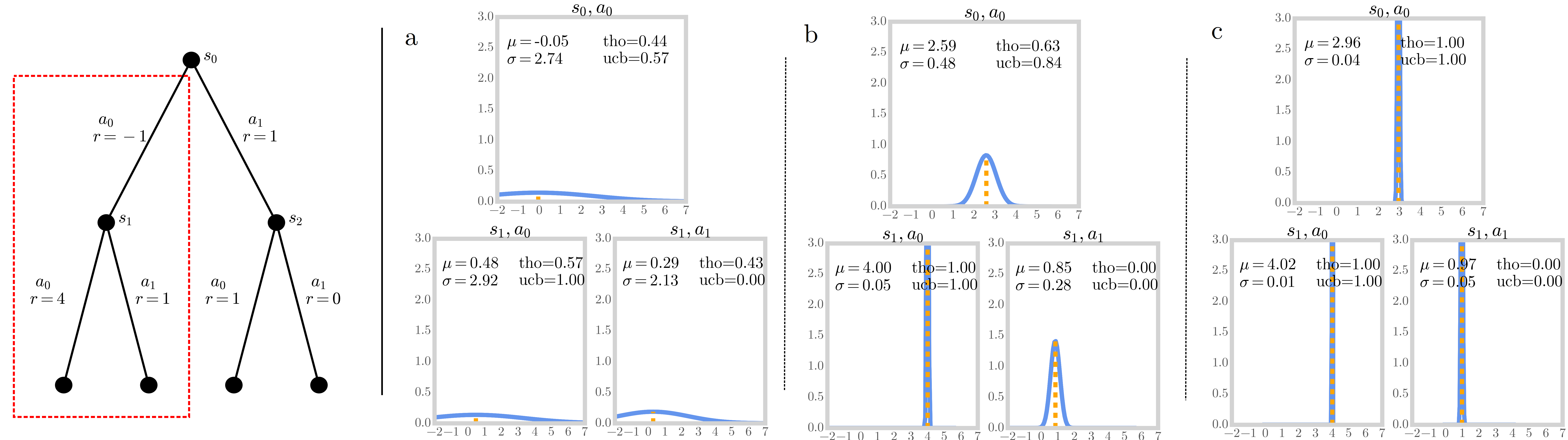}
  \vspace{-0.2cm}
  \caption{\small Gaussian distribution propagation. Left: Example 2-step MDP. Learning process for the state-action pairs in the dotted box is shown on the right. Right: Three (a-c) return exploration phases for the left half of the MDP. Each plot also displays the mean ($\mu$), standard deviation ($\sigma$), and policy probabilities under Thompson (tho) sampling and UCB (ucb).}
    \label{fig_toy_gaussian}
\end{figure*} 

\paragraph{Loss} Next, we want to move our current network predictions $p_\phi(Z|s,a)$ closer to the new target $q(Z|s,a)$, for which we will use a distributional distance. A well-known choice in machine learning is the \textit{cross-entropy} $H(q,p)$:

\begin{equation}
\mathrm{L}_{\text{CE}}(\phi) = H(q(Z),p_\phi(Z)) = \mathrm{E}_{q(Z)} \Big[- \log p_\phi(Z) \Big]. \label{eq_ce}
\end{equation}

For both the Gaussian and Categorical output distributions we can derive closed-form expressions for the cross-entropy $H$, see Appendix \ref{app_losses}. However, for the Gaussian mixture outcome we do not have a closed-form cross-entropy expression, and we instead minimize the $L_2$ distance. See Appendix \ref{app_losses} for details as well. 

In practice, we store a database of transition tuples $\{s,a,r,s',a'\}$, where $a'$ can also be computed either on- or off-policy, and minimize:

\begin{align}
\mathrm{L}_{\text{CE}}(\phi) = \mathrm{E}_{\{s,a,r,s',a'\}\in \mathcal{D}} \Bigg[  \mathrm{E}_{q(Z|s,a)}[ - \log p_\phi(Z|s,a)] \Bigg]
\end{align}

where $q(Z) = f(r,s',a')$ is computed based on Bellman propagation. This completes the policy evaluation step for the Gaussian case.

\section{Distributional Exploration (Policy Improvement)}
Our real interest is usually not in policy evaluation only, as we want to gradually improve our policy as well. The major benefit of probabilistic policy evaluation (previous section) is that we have additional information to balance exploration and exploitation. We will treat the return distribution as something against which we can act optimistically. Exploration under uncertainty has been extensively studied in the bandit literature. Two of the most successful algorithms, which we both consider in this work, are
\begin{enumerate}
\item Thompson sampling \citep{thompson1933likelihood}, which takes a sample $z_a \sim p(Z|s,a)$ for each action and picks the action with the highest draw. 
\item Upper Confidence Bounds (UCB) \cite{auer2002finite}, which picks the action with the highest upper confidence bound $\mu_Z + c \cdot \sigma_Z$ for some constant $c \in \mathbb{R}^+$. Analytic expressions for $\sigma$ for the different output distributions are provided in Appendix \ref{app_sd}.
\end{enumerate}

\section{Experiments} \label{results}
\begin{figure*}[t]
  \centering
      \includegraphics[width = 0.95\textwidth]{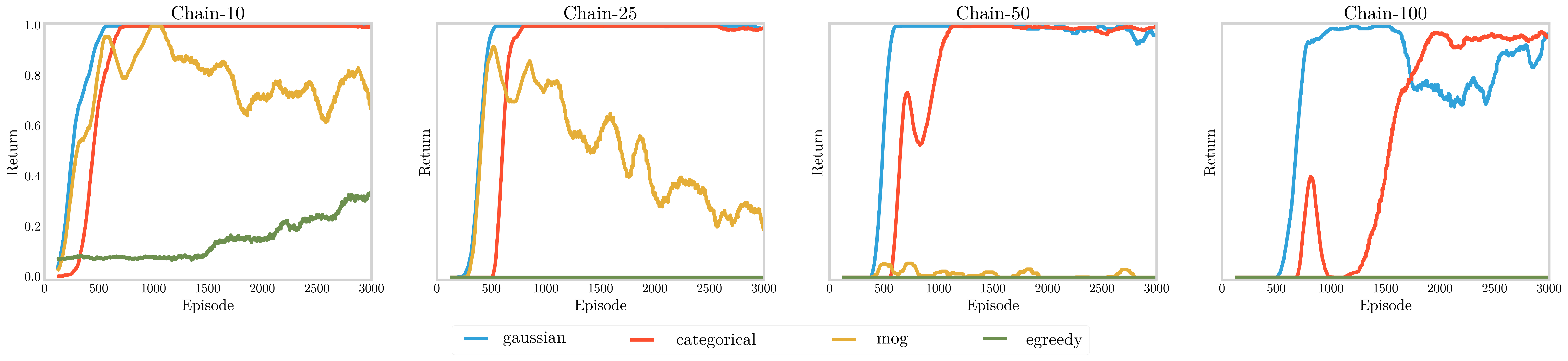}
      \vspace{-0.3cm}
  \caption{\small Learning curves on Chain domain for different types of return distributions. Plots progress row-wise for increased depth of the Chain, i.e. increased exploration difficulty. Exploration uses a UCB policy with constant $c_a \sim \text{Uniform}(1.7,2.3)$ for each $a$ (this induces slight randomness in the otherwise deterministic UCB decision). Results averaged over 10 repetitions.}
    \label{fig_results_chain}
\end{figure*} 

We now show several results of return-based exploration on a Toy example, Chain domain and OpenAI Gym task. Fig. \ref{fig_toy_gaussian}, left shows an example 2-step MDP to illustrate the concept of return-based exploration. On the right of Fig. \ref{fig_toy_gaussian} we display three phases of learning in this MDP for a Gaussian $p_\phi(Z)$. Due to space constraints we only visualize the distributions for the left half of the MDP, the full learning process is shown in Figure \ref{fig_toy_illustration} (Appendix). In Fig. \ref{fig_toy_gaussian}a we just initialized the network, and both Thompson sampling and UCB follow almost uniform policies. After some training (Fig. \ref{fig_toy_gaussian}b) the second state ($s_1$) distributions start converging, but the uncertainty at $s_0$ still remains broader (as it generalizes over the sometimes explored inferior $a_1$ in $s_1$). Thompson sampling and UCB gradually start to prefer $a_0$ in the root state $s_0$ now. Finally, after some additional episodes (Fig. \ref{fig_toy_gaussian}c) the distribution estimates have converged on the optimal state-action values, and both Thompson sampling and UCB have automatically converged on the optimal policy.

We next consider the Chain domain (Appendix \ref{chain}, Figure \ref{chainfigure}), which has been previously studied in RL literature \citep{osband2016deep}. The domain consists of a chain of states of length $N$, with two available actions at each state. The only trace giving a positive, non-zero reward is to select the `correct' action at every step, which is randomly picked at domain initialization. The domain has a strong exploration challenge, which grows exponentially with the length of the chain for undirected exploration methods (see Appendix \ref{chain}).

Figure \ref{fig_results_chain} show the learning curves of return-based exploration for different types of output distributions, and compares the results to $\epsilon$-greedy exploration. The plots progress row-wise to longer chain lengths. First of all, we note that $\epsilon$-greedy learns slowly in the short chain, and does not learn at all in the longer chains. However, the methods with return uncertainty do learn, and consistently solve the domain even for the long chain of length 100. Note that this is a very challenging exploration problem, as we need to take 100 steps correctly while there is no structure in the domain at all (i.e., the correct action randomly changes at each depth in the chain, so a function approximator with local generalization is more harmful than beneficial). The mixture of Gaussians (mog) return distribution performs less stable than the Gaussian and categorical. This might be due to the $L_2$-loss used for the Gaussian mixture, which is different from the cross-entropy losses used for the Gaussian and categorical distributional loss (see Appendix \ref{distributional_details}). In Figure \ref{fig_chain_illustration_appendix} we provide a full illustration of how exploration based on a Categorical return d A detailed illustration of the learning process for the categorical return distribution is shown in Figure \ref{fig_chain_illustration_appendix} (Appendix).

\begin{figure}[t]
  \centering
      \includegraphics[width = 0.27\textwidth]{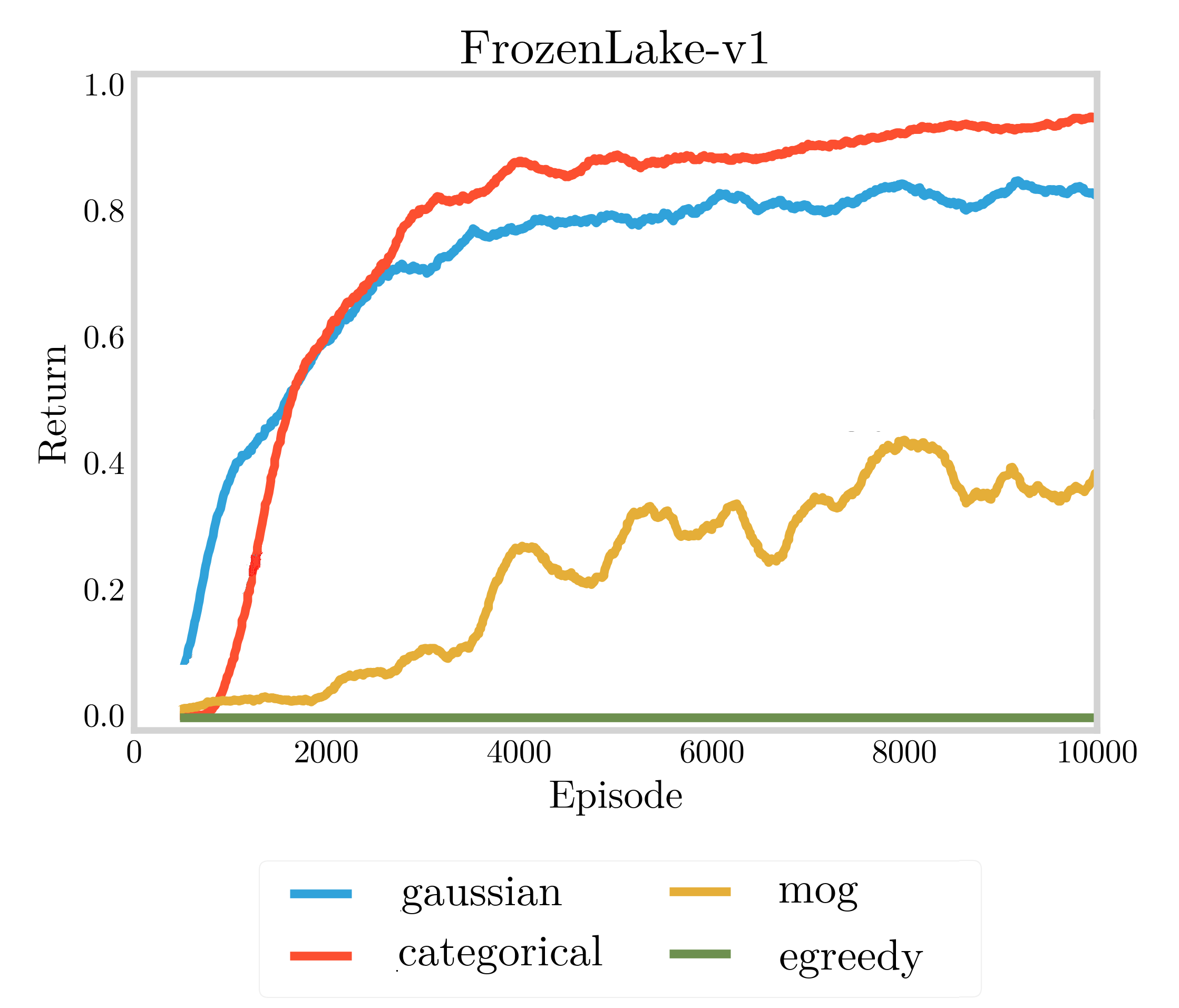}
       \vspace{-0.3cm}
  \caption{\small Return-based exploration versus $\epsilon$-greedy on FrozenLake-v1. The return-based exploration methods use Thompson sampling. Compared to the OpenAI Gym implementation, we modify the environment to be fully deterministic (by removing the random `slipping' effect of the task). Results averaged over 5 repetitions.}
    \label{fig_thompson_detail}
    \vspace{-0.3cm}
\end{figure}

In Figure \ref{fig_thompson_detail} we show the results of return-based exploration on a task from the OpenAI Gym repository: FrozenLake. Again, we observe that the return based exploration methods learn better than $\epsilon$-greedy. These experiments use Thompson sampling for exploration, which shows that return-based exploration can be employed with both UCB or Thompson exploration.

\section{Discussion} \label{future}
We shortly discuss why return-based exploration seems to work so well in the challenging exploration task of the Chain. It turns out that the return distributions really narrow when a certain action terminates the episode. In such cases, we bootstrap a very narrow next state distribution around 0 (because the reward function is assumed deterministic). On the Chain, we see that all the terminating actions very quickly narrow, while the trace along the full path keeps some additional uncertainty. It appears as if the return distribution in this implementation identifies a specific type of uncertainty related to the termination probabilities and asymmetry in the domain search tree, which relates our work to ideas from Monte Carlo Tree Search \citep{moerland2018monte} as well. 

The benefit of exploration based on uncertainty is that policy improvement almost comes for free. The only thing that we propagate are full distributions, which we initialize wide, and then gradually converge when the distributions behind it start converging. This creates a more natural transition from the exploration to the exploitation phase, a trait which most undirected methods ($\epsilon$-greedy, Boltzmann) lack. 

An important direction for future work is to connect the policy-dependent return uncertainty, as studied in this paper, to the {\it statistical} (or epistemic) uncertainty of the mean action-value, which is a function of the local number of visits to a state. The return distribution mechanism in this paper clearly identifies a different aspect of (future policy) uncertainty, which may be related to the termination structure of subtrees, or to the fact that uncertainty in an MDP should propagate over steps as well \citep{dearden1998bayesian}. In any case, due to the sequential nature of MDPs there appear to be more aspects to epistemic/reducible \citep{osband2018randomized} uncertainty, and these distinctions have yet to be properly identified. Finally, another important extension is to stochastic environments \citep{depeweg2016learning,moerland2017learning}, i.e.,  separating which part of the return distribution originates from our own policy uncertainty (for which we can be optimistic) and which part originates from the stochastic environment (for which we want to act on the expectation).

\section{Conclusion}
This paper identified the potential of the return distribution for targeted exploration. In deterministic domains, the return distribution is induced by our own policy, and since we may modify this policy ourselves, it makes sense to act optimistically with respect to this distribution. Exploration based on the return distribution, especially for the Gaussian and Categorical case, manages to solve the `randomized' Chain of length 100 with function approximation, which we believe has not been reported before. Moreover, it also performs well in another task from the OpenAI Gym. Future work should expand these ideas to stochastic environments, and identify the connections to exploration based on the statistical uncertainty of the mean action value. 

\clearpage

\bibliographystyle{icml2018}
\bibliography{uncertainty}

\begin{thebibliography}{30}
\providecommand{\natexlab}[1]{#1}
\providecommand{\url}[1]{\texttt{#1}}
\expandafter\ifx\csname urlstyle\endcsname\relax
  \providecommand{\doi}[1]{doi: #1}\else
  \providecommand{\doi}{doi: \begingroup \urlstyle{rm}\Url}\fi

\bibitem[Auer et~al.(2002)Auer, Cesa-Bianchi, and Fischer]{auer2002finite}
Auer, Peter, Cesa-Bianchi, Nicolo, and Fischer, Paul.
\newblock {Finite-time analysis of the multiarmed bandit problem}.
\newblock \emph{Machine learning}, 47\penalty0 (2-3):\penalty0 235--256, 2002.

\bibitem[Azizzadenesheli et~al.(2017)Azizzadenesheli, Brunskill, and
  Anandkumar]{kamyar2017efficient}
Azizzadenesheli, Kamyar, Brunskill, Emma, and Anandkumar, Animashree.
\newblock {Efficient Exploration through Bayesian Deep Q-Networks}.
\newblock In \emph{{Deep Reinforcement Learning Symposium, NIPS}}, 2017.

\bibitem[Bellemare et~al.(2016)Bellemare, Srinivasan, Ostrovski, Schaul,
  Saxton, and Munos]{bellemare2016unifying}
Bellemare, Marc, Srinivasan, Sriram, Ostrovski, Georg, Schaul, Tom, Saxton,
  David, and Munos, Remi.
\newblock {Unifying count-based exploration and intrinsic motivation}.
\newblock In \emph{{Advances in Neural Information Processing Systems}}, pp.\
  1471--1479, 2016.

\bibitem[Bellemare et~al.(2017)Bellemare, Dabney, and
  Munos]{bellemare2017distributional}
Bellemare, Marc~G, Dabney, Will, and Munos, R{\'e}mi.
\newblock {A distributional perspective on reinforcement learning}.
\newblock \emph{arXiv preprint arXiv:1707.06887}, 2017.

\bibitem[Dearden et~al.(1998)Dearden, Friedman, and
  Russell]{dearden1998bayesian}
Dearden, Richard, Friedman, Nir, and Russell, Stuart.
\newblock {Bayesian Q-learning}.
\newblock In \emph{{AAAI/IAAI}}, pp.\  761--768, 1998.

\bibitem[Depeweg et~al.(2016)Depeweg, Hern{\'a}ndez-Lobato, Doshi-Velez, and
  Udluft]{depeweg2016learning}
Depeweg, Stefan, Hern{\'a}ndez-Lobato, Jos{\'e}~Miguel, Doshi-Velez, Finale,
  and Udluft, Steffen.
\newblock {Learning and policy search in stochastic dynamical systems with
  bayesian neural networks}.
\newblock \emph{arXiv preprint arXiv:1605.07127}, 2016.

\bibitem[Engel et~al.(2005)Engel, Mannor, and Meir]{engel2005reinforcement}
Engel, Yaakov, Mannor, Shie, and Meir, Ron.
\newblock {Reinforcement learning with Gaussian processes}.
\newblock In \emph{{Proceedings of the 22nd international conference on Machine
  learning}}, pp.\  201--208. ACM, 2005.

\bibitem[Gal et~al.(2016)Gal, McAllister, and Rasmussen]{gal2016improving}
Gal, Yarin, McAllister, Rowan~Thomas, and Rasmussen, Carl~Edward.
\newblock {Improving PILCO with bayesian neural network dynamics models}.
\newblock In \emph{{Data-Efficient Machine Learning workshop}}, volume 951,
  pp.\  2016, 2016.

\bibitem[Guez et~al.(2012)Guez, Silver, and Dayan]{guez2012efficient}
Guez, Arthur, Silver, David, and Dayan, Peter.
\newblock {Efficient Bayes-adaptive reinforcement learning using sample-based
  search}.
\newblock In \emph{{Advances in Neural Information Processing Systems}}, pp.\
  1025--1033, 2012.

\bibitem[Henderson et~al.(2017)Henderson, Doan, Islam, and
  Meger]{henderson2017bayesian}
Henderson, Peter, Doan, Thang, Islam, Riashat, and Meger, David.
\newblock {Bayesian Policy Gradients via Alpha Divergence Dropout Inference}.
\newblock \emph{arXiv preprint arXiv:1712.02037}, 2017.

\bibitem[Jeong \& Lee(2017)Jeong and Lee]{jeong2017bayesian}
Jeong, Heejin and Lee, Daniel~D.
\newblock {Bayesian Q-learning with Assumed Density Filtering}.
\newblock \emph{arXiv preprint arXiv:1712.03333}, 2017.

\bibitem[Kaufmann \& Koolen(2017)Kaufmann and Koolen]{kaufmann2017monte}
Kaufmann, Emilie and Koolen, Wouter~M.
\newblock {Monte-carlo tree search by best arm identification}.
\newblock In \emph{{Advances in Neural Information Processing Systems}}, pp.\
  4897--4906, 2017.

\bibitem[Mannor \& Tsitsiklis(2011)Mannor and Tsitsiklis]{mannor2011mean}
Mannor, Shie and Tsitsiklis, John.
\newblock {Mean-variance optimization in Markov decision processes}.
\newblock \emph{arXiv preprint arXiv:1104.5601}, 2011.

\bibitem[Moerland et~al.(2017{\natexlab{a}})Moerland, Broekens, and
  Jonker]{moerland2017efficient}
Moerland, Thomas~M, Broekens, Joost, and Jonker, Catholijn~M.
\newblock {Efficient exploration with Double Uncertain Value Networks}.
\newblock \emph{arXiv preprint arXiv:1711.10789}, 2017{\natexlab{a}}.
\newblock Deep Reinforcement Learning Symposium @ NIPS 2017.

\bibitem[Moerland et~al.(2017{\natexlab{b}})Moerland, Broekens, and
  Jonker]{moerland2017learning}
Moerland, Thomas~M, Broekens, Joost, and Jonker, Catholijn~M.
\newblock {Learning Multimodal Transition Dynamics for Model-Based
  Reinforcement Learning}.
\newblock \emph{arXiv preprint arXiv:1705.00470}, 2017{\natexlab{b}}.

\bibitem[Moerland et~al.(2018)Moerland, Broekens, Plaat, and
  Jonker]{moerland2018monte}
Moerland, Thomas~M, Broekens, Joost, Plaat, Aske, and Jonker, Catholijn~M.
\newblock {Monte Carlo Tree Search for Asymmetric Trees}.
\newblock \emph{arXiv preprint arXiv:1805.09218}, 2018.

\bibitem[Morimura et~al.(2012)Morimura, Sugiyama, Kashima, Hachiya, and
  Tanaka]{morimura2012parametric}
Morimura, Tetsuro, Sugiyama, Masashi, Kashima, Hisashi, Hachiya, Hirotaka, and
  Tanaka, Toshiyuki.
\newblock {Parametric return density estimation for reinforcement learning}.
\newblock \emph{arXiv preprint arXiv:1203.3497}, 2012.

\bibitem[Osband et~al.(2014)Osband, {Van Roy}, and
  Wen]{osband2014generalization}
Osband, Ian, {Van Roy}, Benjamin, and Wen, Zheng.
\newblock {Generalization and exploration via randomized value functions}.
\newblock \emph{arXiv preprint arXiv:1402.0635}, 2014.

\bibitem[Osband et~al.(2016)Osband, Blundell, Pritzel, and {Van
  Roy}]{osband2016deep}
Osband, Ian, Blundell, Charles, Pritzel, Alexander, and {Van Roy}, Benjamin.
\newblock {Deep exploration via bootstrapped DQN}.
\newblock In \emph{{Advances in Neural Information Processing Systems}}, pp.\
  4026--4034, 2016.

\bibitem[Osband et~al.(2018)Osband, Aslanides, and
  Cassirer]{osband2018randomized}
Osband, Ian, Aslanides, John, and Cassirer, Albin.
\newblock {Randomized Prior Functions for Deep Reinforcement Learning}.
\newblock \emph{arXiv preprint arXiv:1806.03335}, 2018.

\bibitem[Petersen et~al.(2008)Petersen, Pedersen, et~al.]{petersen2008matrix}
Petersen, Kaare~Brandt, Pedersen, Michael~Syskind, et~al.
\newblock {The matrix cookbook}.
\newblock \emph{Technical University of Denmark}, 7\penalty0 (15):\penalty0
  510, 2008.

\bibitem[Russo et~al.(2017)Russo, {Van Roy}, Kazerouni, and
  Osband]{russo2017tutorial}
Russo, Daniel, {Van Roy}, Benjamin, Kazerouni, Abbas, and Osband, Ian.
\newblock {A Tutorial on Thompson Sampling}.
\newblock \emph{arXiv preprint arXiv:1707.02038}, 2017.

\bibitem[Sobel(1982)]{sobel1982variance}
Sobel, Matthew~J.
\newblock {The variance of discounted Markov decision processes}.
\newblock \emph{Journal of Applied Probability}, 19\penalty0 (4):\penalty0
  794--802, 1982.

\bibitem[Sutton \& Barto(1998)Sutton and Barto]{sutton1998reinforcement}
Sutton, Richard~S and Barto, Andrew~G.
\newblock \emph{{Reinforcement learning: An introduction}}, volume~1.
\newblock MIT press Cambridge, 1998.

\bibitem[Tamar et~al.(2016)Tamar, {Di Castro}, and Mannor]{tamar2016learning}
Tamar, Aviv, {Di Castro}, Dotan, and Mannor, Shie.
\newblock {Learning the variance of the reward-to-go}.
\newblock \emph{Journal of Machine Learning Research}, 17\penalty0
  (13):\penalty0 1--36, 2016.

\bibitem[Tang \& Agrawal(2018)Tang and Agrawal]{tang2018exploration}
Tang, Yunhao and Agrawal, Shipra.
\newblock {Exploration by Distributional Reinforcement Learning}.
\newblock \emph{arXiv preprint arXiv:1805.01907}, 2018.

\bibitem[Tang \& Kucukelbir(2017)Tang and Kucukelbir]{tang2017variational}
Tang, Yunhao and Kucukelbir, Alp.
\newblock {Variational Deep Q Network}.
\newblock \emph{arXiv preprint arXiv:1711.11225}, 2017.

\bibitem[Tesauro et~al.(2012)Tesauro, Rajan, and Segal]{tesauro2012bayesian}
Tesauro, Gerald, Rajan, VT, and Segal, Richard.
\newblock {Bayesian inference in monte-carlo tree search}.
\newblock \emph{arXiv preprint arXiv:1203.3519}, 2012.

\bibitem[Thompson(1933)]{thompson1933likelihood}
Thompson, William~R.
\newblock {On the likelihood that one unknown probability exceeds another in
  view of the evidence of two samples}.
\newblock \emph{Biometrika}, 25\penalty0 (3/4):\penalty0 285--294, 1933.

\bibitem[White(1988)]{white1988mean}
White, DJ.
\newblock {Mean, variance, and probabilistic criteria in finite Markov decision
  processes: a review}.
\newblock \emph{Journal of Optimization Theory and Applications}, 56\penalty0
  (1):\penalty0 1--29, 1988.

\end{thebibliography}

\appendix

\section{Distributional Details} \label{distributional_details}
The current network distributions are denote by $p_\phi(Z|s,a)$, which we want to update with a newly calculated target distribution $q(Z|s,a)$. For readability, we will omit the dependency on $s,a$ in the remainder of this section. We study three types of network output distributions:

\begin{enumerate}
\item Gaussian: $p(Z) = \mathcal{N}(Z|\mu,\sigma)$
\item Categorical: $p(Z)$ parametrized by the number of bins $N \in \mathbb{N}$ and edges $Z_{min},Z_{max} \in \mathbb{R}$. Define the set of bins as $\{z_i = Z_{min} + 0.5\Delta z + i \Delta z : 0 \leq i < N \}$, for $\Delta z := \frac{Z_{max} - Z_{min}}{N}$. Each bin has associated density $p(z_i)$, with $\sum_{i=1}^N p(z_i) = 1$.
\item Gaussian mixture: $p(Z) = \sum_{i=1}^M p_i \mathcal{N}(Z|\mu_i,\sigma_i) $, for $M$ mixtures. Here $p_i$ denotes the weight of the $i$-th mixture component, $\sum_i p_i = 1$.  
\end{enumerate}

We now detail the loss, Bellman propagation and analytic standard deviation (as used in the UCB policy) for each of these output distributions.

\subsection{Loss} \label{app_losses}

\paragraph{Gaussian}
The main text already introduced the cross-entropy loss $\mathrm{L}_{CE} = H(q,p)$ used for Gaussian $p_\phi(Z)$. Here we derive the analytical expression of this cross-entropy:

\begin{align}
H(q,p) &= \mathrm{E}_{q(Z)}[- \log p(Z)] \nonumber \\
&= - \int q(Z) \log \frac{1}{\sqrt{2\pi\sigma_p^2}} \exp(\frac{-(Z-\mu_p)^2}{2\sigma_p^2} ) \mathrm{d} Z.
\end{align}

Bringing everything that does not depend on $Z$ out of the integral and taking the logarithm:
\begin{equation}
H(q,p) = \frac{1}{2} \log (2 \pi \sigma_p^2) + \frac{1}{2\sigma_p^2} \int q(Z) (Z-\mu_p)^2) \mathrm{d} Z
\end{equation}

Which can be rewritten as
\begin{align}
H(q,p) &= \frac{1}{2} \log (2 \pi \sigma_p^2) + \frac{1}{2\sigma_p^2} \Bigg[\mathrm{E}_{q(Z)} [Z^2] \nonumber \\ 
& - \mathrm{E}_{q(Z)} [Z \mu_p] + \mathrm{E}_{q(Z)} [\mu_p^2] \Bigg] \nonumber \\
\end{align}

We can rewrite the second moment $\mathrm{E}_{q(Z)}[Z^2]$ in terms of the mean and variance of $Z$:

\begin{equation}
\mathrm{E}_{q(Z)} [Z^2] = \sigma_q^2 + \mu_q^2. \label{eq_simplify_second_moment}
\end{equation}

Therefore, we can simplify the full expression to

\begin{align}
H(q,p) &= \frac{1}{2} \log (2 \pi \sigma_p^2) + \frac{1}{2\sigma_p^2} (\sigma_q^2 + \mu_q^2 - 2 \mu_q \mu_p + \mu_p^2 ) \nonumber \\ 
&= \frac{1}{2} \log (2 \pi \sigma_p^2) + \frac{\sigma_q^2 + (\mu_q - \mu_p)^2}{2\sigma_p^2}
\end{align}

which is used as the closed-form loss for the Gaussian experiments (Eq. \ref{eq_ce}) in this paper. Note that we also experimented with (other) closed form distributional losses for Gaussians, such as the Bhattacharyya distance and Hellinger distance, but these did not significantly improve performance. 

\paragraph{Categorical}
For the categorical target distribution $q(Z)$ we again minimize the cross-entropy $H$ with $p_\phi(Z)$: 

\begin{equation}
\mathrm{L}_{CE} = H(q(Z),p_\phi(Z)) = - \sum_i q(Z_i) \log p_\phi(Z_i).
\end{equation}

\paragraph{Gaussian Mixture}
There is no closed form expression for the KL-divergence or cross-entropy between two Gaussian mixtures. We could of course approximate such a loss by repeated sampling, but this will strongly increase the computational burden. Therefore, we instead searched for a distance measure between Gaussian mixtures that does have a closed form expression, which is the $L_2$-distance:

\begin{align}
\mathrm{L}_{L2}(q(Z),p_\phi(Z)) &= \int \Big( q(Z) - p(Z;\phi) \Big)^2 \mathrm{d} Z \nonumber \\ 
&= \sum_{i,i'} q_i q_{i'} \int q_i(Z) q_{i'}(Z) \mathrm{d} Z  \nonumber \\
& + \sum_{j,j'} p_j p_{j'} \int p_i(Z) p_{i'}(Z) \mathrm{d} Z \nonumber \\ 
& - 2 \sum_{i,j} q_i p_j \int q_i(Z) p_j(Z) \mathrm{d} Z. \label{eq_mog1}
\end{align} 

We may simplify the remaining integrals in this expression, since for any two Gaussians $\mathcal{N}_1$ and $\mathcal{N}_2$ we have \citep{petersen2008matrix}:

\begin{equation}
\int \mathcal{N}_1(Z|\mu_1,\sigma_1) \mathcal{N}_2(Z|\mu_2,\sigma_2) \mathrm{d} Z = \mathcal{N}(\mu_1|\mu_2,\sigma_1 + \sigma_2),
\end{equation}

Therefore, Eq. \ref{eq_mog1} simplies to

\begin{align}
\mathrm{L}_{L2}(q(Z),p_\phi(Z)) &= \sum_{i,i'} q_i q_{i'} \mathcal{N}(\mu^q_{i}|\mu^q_{i'},\sigma^q_i + \sigma^q_{i'})  \nonumber \\
& + \sum_{j,j'} p_j p_{j'} \mathcal{N}(\mu^p_j|\mu^p_{j'},\sigma^p_j + \sigma^p_{j'}) \nonumber \\ 
& - 2 \sum_{i,j} q_i p_j \mathcal{N}(\mu^q_i|\mu^p_j,\sigma^q_i + \sigma^p_j), 
\end{align} 

which can be evaluated in $O(M^2)$ time for $M$ mixture components. 

\paragraph{Sample-based loss}
For some output distributions we either do not have a density (like some deep generative models) or the available analytic distributional loss performs suboptimal. However, we can always sample from our model. For example, for a 1-step Q-learning update, we can (repeatedly) sample from our network at the next timestep $Z_k' \sim p(Z|s',a')$, transform these through the Bellman equation, and then train our model on a negative log-likelihood loss:

\begin{align} 
\mathrm{L}_{NLL} &= \mathbb{E}_{q(Z)} [- \log p_\phi(Z) ] \nonumber \\
 &\approx - \sum_k  \log p_\phi(r + \gamma \cdot Z_k'), \quad Z_k' \sim p(Z|s',a')
\end{align}

Results of this approach are not shown, but were comparable to the results with approximate return propagation shown in the Results section. However, this approach is clearly more computationally expensive. 

\subsection{Bellman Propagation} \label{app_bellman}

Given a data tuple $\{s,a,r,s',a'\}$, where $a'$ may either be on- or off-policy, and a bootstrapped distribution $p_\phi(z'|s',a')$, we want to calculate the one-step Bellman transformed distribution $\mathcal{T} p(z|s,a)$.

\paragraph{Categorical}
For the categorical distribution, we may Bellman transform each individual atom/bin, and then project the probabilities of the transformed means back on the atoms (denoted by operator $\Psi$). This procedure follows \citet{bellemare2017distributional}: 

\begin{align}
q(Z_i) &= \Big(\Psi \mathcal{T} Z \Big)_i \nonumber \\
&= \sum_{j=1}^{N} p_{\phi^\dagger}(Z'_j) \cdot \Big[ 1 - \frac{[r + \gamma Z'_j]_{Z_{min}}^{Z_{max}} - Z_i}{\Delta Z}  \Big]_0^1 
\end{align}

\paragraph{Gaussian mixture}
For the Gaussian mixture case, we have

\begin{align}
q(Z) &= \mathcal{T} Z(s,a) \nonumber \\
& = r(s,a) + \gamma \mathrm{E}_{a' \sim \pi(\cdot|s')} [Z'] \nonumber \\
&= r(s,a) + \gamma \mathrm{E}_{a' \sim \pi(\cdot|s')} [ \sum_i p_i Z'_i ] \nonumber \\
&= \sum_i p_i \Big[ r(s,a) + \gamma \mathrm{E}_{a' \sim \pi(\cdot|s')} [Z_i] \Big] 
\end{align}

This implies that we may propagate each Gaussian mixture component individually, as discussed in Section \ref{return_propagation}, keeping each mixture weight the same. 

\subsection{Standard Deviation} \label{app_sd}
For UCB exploration, we require fast (i.e., analytic) access to the distribution standard deviation, to prevent repeatedly having to sample. Clearly, for the Gaussian output we directly have the standard deviation available. 

\paragraph{Categorical}
For a categorical output distribution with categories $z_i$ and associated probabilities $p(z_i)$, we have the standard deviation as:

\begin{equation}
\mathrm{Sd}[Z] = \sum_i p(Z_i) (Z_i - \mathrm{E}[Z])^2
\end{equation} 

where $\mathrm{E}[Z] = \sum_i Z_i \cdot p(Z_i) $. 

\paragraph{Gaussian mixture}
For a Gaussian mixture model with mixture weights $p_i$, mixture means $\mu_i$ and mixture standard deviation $\sigma_i$, we start from:

\begin{align}
\mathrm{Var}[y] &= \mathrm{E}_{p(Z)}[Z^2] - (\mathrm{E}_{p(Z)}[Z])^2 \nonumber \\
&= \sum_i p_i \cdot \mathrm{E}_{p(Z_i)}[Z_i^2] - (\sum_i p_i \cdot \mu_i)^2 \label{eq13}
\end{align} 

Now we may again use Eq. \ref{eq_simplify_second_moment} to rewrite the second moments of the mixture components in terms of their means and variances, i.e. $\mathrm{E}_{p(Z_i)}[Z_i^2] = \sigma_i^2 + (\mu_i)^2$. Plugging this expression into Eq. \ref{eq13} gives:

\begin{align}
\mathrm{Var}[y] &= \sum_i p_i \cdot \Big( \sigma_i^2 + (\mu_i)^2 \Big) - (\sum_i p_i \cdot \mu_i)^2 \nonumber \\
&= \sum_i p_i \cdot \sigma_i^2 + \sum_i p_i (\mu_i)^2 - (\sum_i p_i \cdot \mu_i)^2
\end{align} 

This last expression gives the variance of the mixture in terms of the weight, mean and variance of the mixture components.

%%%%%%%%%

\section{Related Work} \label{relatedwork}

\paragraph{Return Uncertainty}
While the distributional Bellman equation (Eq. \ref{eq_distr_bellman}) is certainly not new \cite{sobel1982variance,white1988mean}, nearly all RL research has focussed on the mean action-value. Most papers that do study the underlying return distribution study the 'variance of the return'. \citet{engel2005reinforcement} learned the distribution of the return with Gaussian Processes, but did not use it for exploration. \citet{tamar2016learning} studied the variance of the return with linear function approximation. \citet{mannor2011mean} theoretically studies policies that bound the variance of the return. 

The variance of the return has actually primarily been in the context of {\it risk-sensitive RL}. In several scenarios we may want to avoid incidental large negative pay-offs, which can e.g. be disastrous for a real-world robot, or in a financial portfolio. \citet{morimura2012parametric} studied parametric return distribution propagation as well. They do risk-sensitive exploration by softmax exploration over {\it quantile} Q-functions (also known as the {\it Value-at-Risk} (VaR) in financial management literature). Their distribution losses are based on KL-divergences (including Normal, Laplace and skewed Laplace distributions), but their implementations do remain in the tabular setting.

\citet{bellemare2017distributional} was the first to theoretically study the distributional Bellman operator, and also implement a distributional policy evaluation algorithm in the context of neural networks. Thereby, there work can be considered the basis of our work, where we present an extension that uses the return distribution for exploration. Concurrently with our work, \citet{tang2018exploration,tang2017variational} interpreted the return distribution from a variational perspective and leveraged it for exploration as well. \citet{moerland2017efficient} also provided initial work on the return distribution for exploration. Our present paper is more extensive on the theoretical side, for example specifying full distributional loss functions and comparing different types of network output distributions. However, \citet{moerland2017efficient} does try to connect the concept of return distribution to the statistical uncertainty of the mean action value as well, which both seem plausible quantities for exploration.

Another branch of related work is from the Tree Search community. Various papers have focussed on propagating distributions within the tree, e.g. \citet{tesauro2012bayesian} and \citet{kaufmann2017monte}. The tree search approach by \citet{moerland2018monte} does not explicitly propagate distributions (only $\sigma$-like estimates), but their idea (the remaining uncertainty should also incorporate the remaining uncertainty in the subtree below an action) is observable in the return-based exploration and learning visualizations of this paper as well.

\paragraph{Other Uncertainty-based exploration methods}
There exists a long history of work on the statistical uncertainty of the mean action value for exploration, in the context of function approximation for example by \citet{osband2016deep}, \citet{gal2016improving} and more recently \citet{kamyar2017efficient}, \citet{henderson2017bayesian} and \citet{jeong2017bayesian}. Moreover, the uncertainty theme for exploration also appears in count-based exploration approaches \citep{bellemare2016unifying} and model-based RL \cite{guez2012efficient,moerland2017learning}. 

\section{Randomized Chain} \label{chain}

\begin{figure}[ht]
  \centering
      \includegraphics[width = 0.45\textwidth]{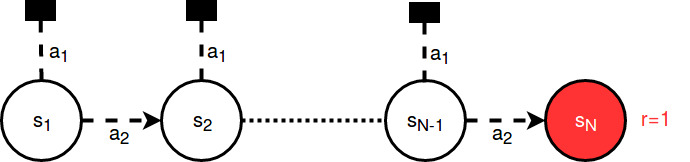}
  \caption{\small Chain domain. Example MDP where undirected exploration is highly inefficient. Based on \citet{osband2014generalization}.}
    \label{chainfigure}
\end{figure}

We here present the randomized Chain, which we believe is the correct implementation of a well-known RL task known as the Chain \citep{osband2014generalization} (Fig. \ref{chainfigure}). The domain illustrates the difficulty of exploration with sparse rewards.The MDP consists of a chain of states $\mathcal{S} = \{1,2...,N\}$. At each time step the agent has two available actions: $a_1$ (`left') and $a_2$ (`right'). At every step, one of both actions is the `correct' one, which deterministically moves the agent one step further in the chain. The wrong action terminates the episode. All states have zero reward except the final chain state $N$, which has $r=1$. 

Variants of these problem have been studied more frequently in RL \citep{osband2014generalization}. In the `ordered' implementation, the correct action is always the same (e.g. $a_2$), and the optimal policy is to always walk right. This is the variant illustrated in Fig. \ref{chainfigure} as well. However, in our `randomized' Chain implementation the correct action is randomly picked at domain initialization. The problem with the ordered version is that it introduced a systematic bias which is easily exploited when learning with neural networks. Due to the generalization of neural networks, it relatively easily predicts to always take action $a_2$, and then suddenly solves the entire chain. With the randomized version, there is actually no structure in the domain at all, and learning with a neural network only makes the domain more complicated. The `randomized' version therefore gives the true exponential complexity, as reported before \citep{osband2014generalization,moerland2017efficient}, when learning with neural networks.

\section{Implementation Details} \label{appendix_implementation}
Network architecture consists of a 3 layer neural network {\it per discrete action} with 256 nodes in each hidden layer and ELU activations. Learning rates were 0.0005 on all experiments. Optimization is performed with stochastic gradient descent on minibatches of size 32 using Adam updates in Tensorflow. We use a replay database of size 50.000. After collecting a new (set of) roll-outs, we randomly sample an equal amount of data from the replay for processing. All new collected data is processed on-policy, while all replay data is processed off-policy. The maximum length per episode is 200. We use discount factor $\gamma = 0.995$. All $\epsilon$-greedy experiments have $\epsilon$ fixed at 0.05 throughout learning.

For the categorical outcome we put the bin edges slightly above and below the highest and lowest expected reward in the domain. In the chain we use $N=7$ bins, on the other domains we use $N=31$ bins. For the Gaussian output we add an initialization bias of $1$ to the standard deviation at initialization. For the Gaussian mixture output we use $M=5$ mixture components, where we spread out the mixture means upon initialization. Due to the logarithm appearing in the Gaussian cross-entropy loss we see that the gradient may explode when the standard deviation strongly narrows. We mitigate this problem by clipping gradients. 

Thompson sampling is best implemented in an `episode-wise' fashion, where we sample from a posterior distribution over parameters once at the beginning of a new episode \cite{russo2017tutorial}. This ensures deep exploration. However, for the return based uncertainty we directly sample in the network output space per action, and we cannot implement this correlated form of Thompson sampling. 

Full code is available from \url{https://github.com/tmoer/return_distribution_exploration.git}. 

\begin{figure*}[ht]
  \centering
      \includegraphics[height = 0.8\textheight]{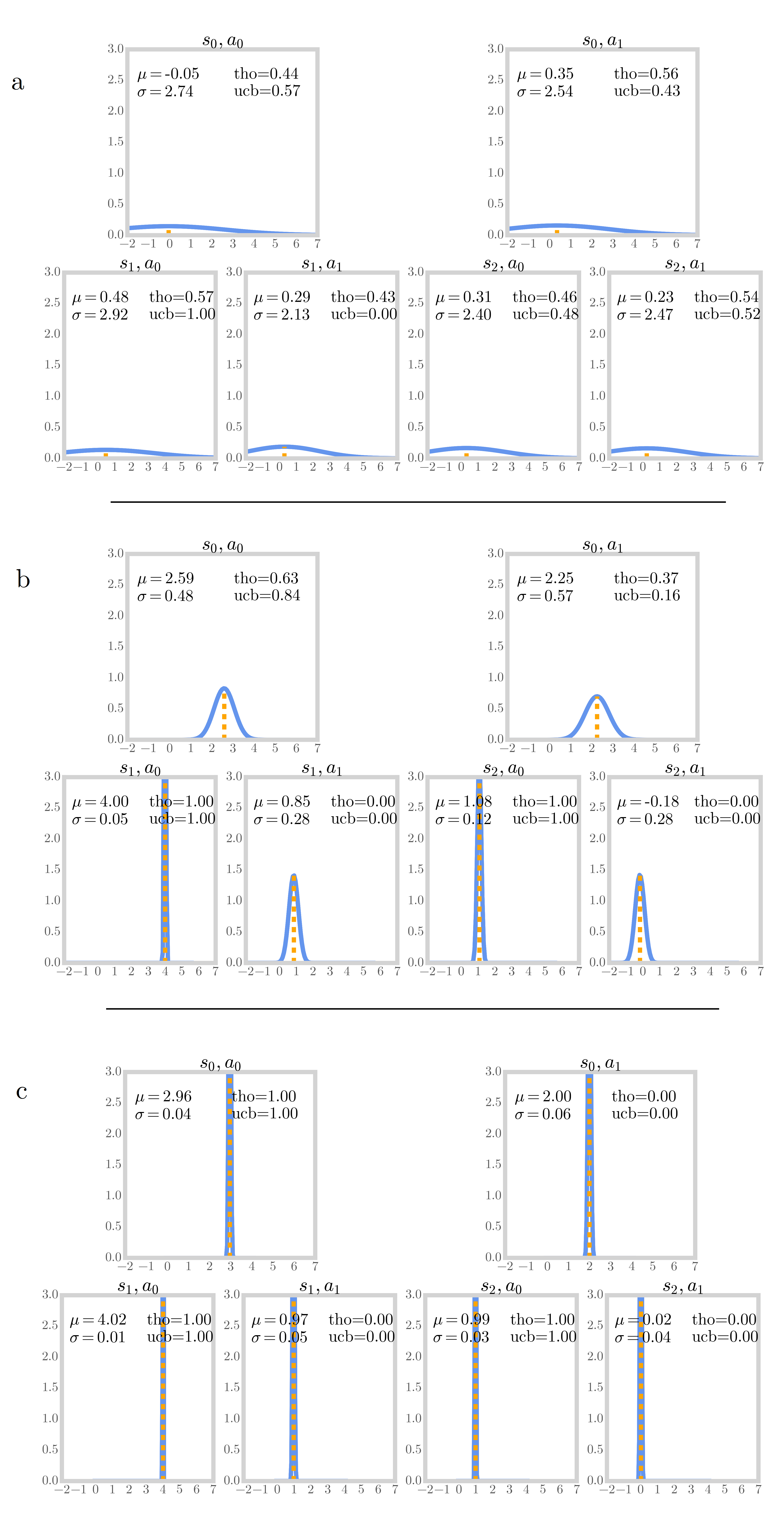}
  \caption{\small Example of Gaussian distribution propagation on the Toy example of Fig.\ref{fig_toy_gaussian}. {\bf a)} Return distributions at initialization. Both Thompson sampling and UCB have largely uniform policies. {\bf b)} Distributions after training for 64 episodes. The terminal state distributions start gradually converging, while the distributions at state $s_0$ remain broader. The terminal node decisions are already greedy, while the first node already starts to assign higher probability to the optimal action $a_0$. {\bf c)} Converged distributions after training for some additional time. Both UCB and Thompson sampling now deterministically sample the optimal policy.}
    \label{fig_toy_illustration}
\end{figure*}

\begin{figure*}[ht]
  \centering
      \includegraphics[width = 1.0\textwidth]{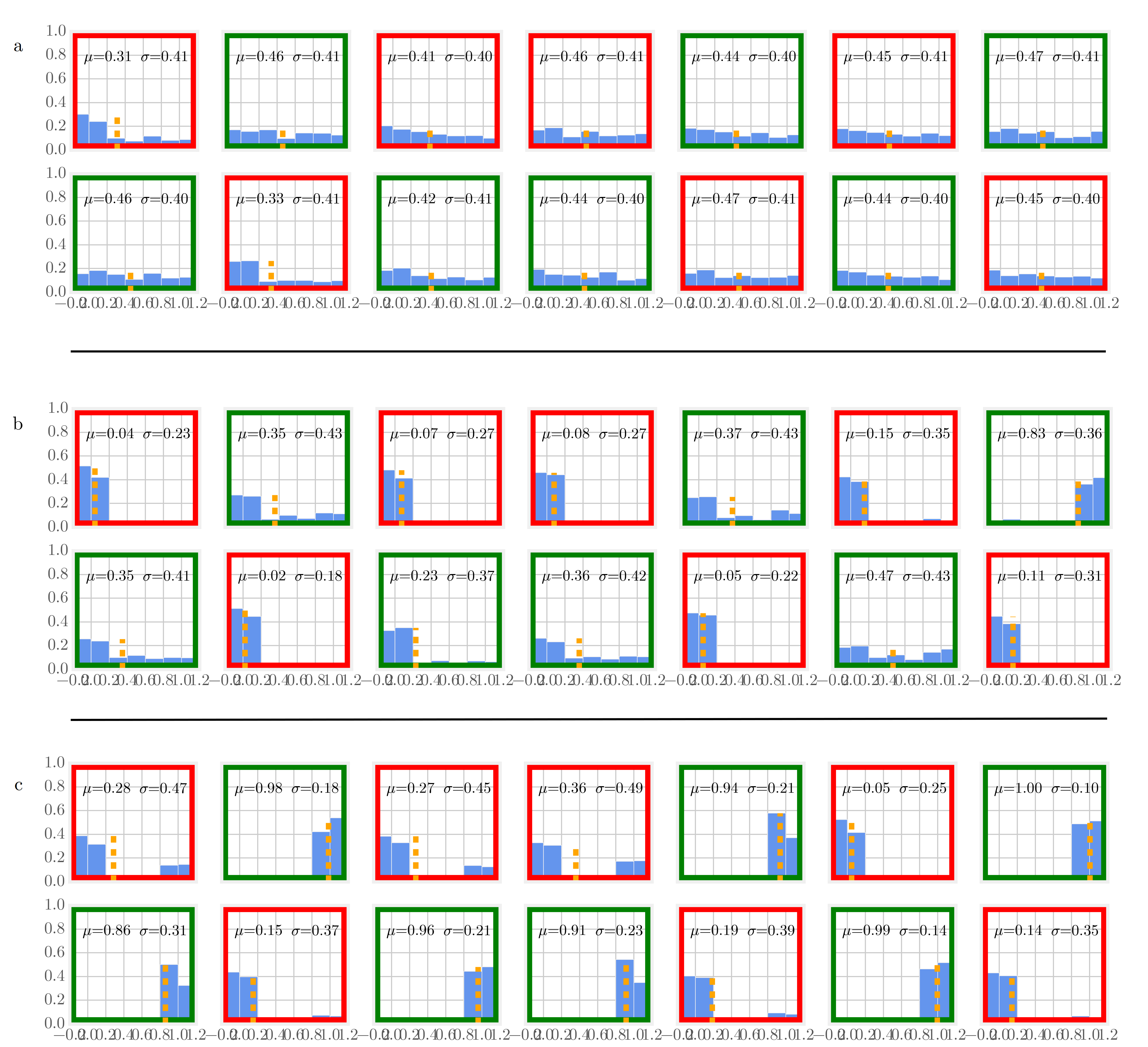}
  \caption{\small Example of return distribution-based exploration on the Chain of length 7. Each plot (a-c) shows successive states (left-to-right) and both actions (up and down). The correct action at each step (randomly drawn at domain initialization) is indicated by a green box around the plot. We use a categorical $p(Z)$ with 7 atoms between $-0.2$ and $1.2$. {\bf a)}. Return distributions after 2 episodes. The distributions are almost uniform, which makes the policy fully exploratory. {\bf b)}. Return distributions after 28 episodes. Both the correct and wrong action have propagated mass towards 0. However, the distributions of the wrong actions converge faster, because the correct actions propagate the remaining uncertainty at the next timestep. The correct action in the last state already started to move towards a value of 1. {\bf c)}. Converged return distributions after 68 episodes. All the correct actions have now backpropagated the return from the end of the chain. The policy now consistently exploits. Note that some of the wrong actions (red boxes) put some mass at a return of 1 as well. This is due to the generalization from neighboring states (which are treated as continuous in the network input), but disappears with enough data.}
    \label{fig_chain_illustration_appendix}
\end{figure*} 
\clearpage
\end{document}